\documentclass[11pt]{article}

\usepackage[preprint]{acl}

\usepackage{times}
\usepackage{latexsym}

\usepackage[T1]{fontenc}

\usepackage[utf8]{inputenc}

\usepackage{microtype}

\usepackage{inconsolata}

\usepackage{graphicx}

\usepackage{hyperref}
\hypersetup{colorlinks=true}

\usepackage{amsmath}
\usepackage{amsfonts}
\usepackage{bbm}
\usepackage{accents} 

\usepackage{booktabs}
\usepackage{multirow}
\usepackage{hhline} 
\usepackage{makecell} 

\makeatletter
\AtBeginDocument{%
  \let\ACL@orig@maketitle\@maketitle
  \def\@maketitle{%
    \ACL@orig@maketitle
    \ifx\@date\@empty\else
      \vspace{-0.5em}%
      \begin{center}
        \begin{minipage}{0.97\textwidth}
          \normalsize
          \@date
        \end{minipage}
      \end{center}%
      \vspace{1em}%
    \fi
  }%
}
\makeatother

\title{Building AI Agents to Improve Job Referral Requests to Strangers}

\author{Ross Chu \\
  \\
  \texttt{ross.chu@berkeley.edu} \And
  Yuting Huang \\
  \\
  \texttt{yuting\_huang@berkeley.edu} }

\date{%
    \vspace{-2em}
    \centerline{December 2025\thanks{
    Ross Chu acknowledges financial support through the National Science Foundation Graduate Research Fellowship under Grant DGE 2146752. This work does not represent the views of NSF or UC Berkeley. This paper builds on a prior project titled ``Predicting the Success of Job Referral Requests to Strangers'', which was submitted for a course at UC Berkeley. 
}}%
    \vspace{2em}
    \centerline{\textbf{Abstract}}%
    \vspace{1em}
    This paper develops AI agents that help job seekers write effective requests for job referrals in a professional online community. 
    The basic workflow consists of an improver agent that rewrites the referral request 
    and an evaluator agent that measures the quality of revisions using a model trained to predict the probability of receiving referrals from other users. 
    Revisions suggested by the LLM (large language model) increase predicted success rates for weaker requests while reducing them for stronger requests. 
    Enhancing the LLM with Retrieval-Augmented Generation (RAG) prevents edits that worsen stronger requests while it amplifies improvements for weaker requests. Overall, using LLM revisions with RAG increases the predicted success rate for weaker requests by 14\% without degrading performance on stronger requests.  
    Although improvements in model-predicted success do not guarantee more referrals in the real world, they provide low-cost signals for promising features before running higher-stakes experiments on real users.
\\
}

\begin{document}
\maketitle

\section{Introduction}

    For job applicants, getting employee referrals can significantly increase their chances of securing interviews. Employee referrals not only increase the chances of getting an interview, but also enhance the overall quality of employees who are ultimately hired \citep{ilyes2023, hensvik2016}. However, not everyone has the same level of access to professional networks, and many job seekers rely on online communities to seek employee referrals from strangers. 
    
    This paper develops AI agents to help job seekers write effective requests for job referrals. Our context is the ``Jobs \& Referrals'' channel on \hyperlink{teamblind.com}{teamblind.com}, where job seekers connect with current employees who are interested in offering referrals. Crafting an effective request is not a straightforward task: employees are selective about giving referrals, and 54\% of requests fail to attract any referrals at all. This tool can help job seekers by increasing their chances of receiving referrals from the online community. 
    
    In the basic workflow, an improver agent rewrites the referral request by making API calls to the LLM (large language model), and an evaluator agent compares the quality of referral requests before VS after LLM revisions. The quality of a referral request is measured by a model trained to predict its probability of receiving referrals from other users. We consider three models for this prediction task: 1) a sentence transformer that converts text into vector embeddings, 2) a TF-IDF model with scaled frequencies for unigrams and bigrams, and 3) a linguistic model with a rich set of semantic features motivated by our domain knowledge of the platform and prior literature on making effective requests. The best performing model is a sentence transformer fine-tuned to achieve an AUROC of 0.681, which is comparable to altruistic requests in \citet{althoff2014}. 
    
    We enhance the basic workflow with Retrieval-Augmented Generation (RAG). In this workflow, the retriever agent selects well-written examples for the LLM that are contextually relevant to the user's request. The explainer agent gives editorial guidance by providing ratings on the overall request, its title, and each individual sentence. These ratings can take on one of three values (``strong'', ``weak'', or ``moderate''), which guide the LLM on how heavily to edit the request and which segments it should focus on. 

    Revisions suggested by the LLM have asymmetric effects on the quality of referral requests, which depends on their initial probability of receiving referrals. The basic workflow increases predicted success rates for weaker requests, but it reduces them for stronger requests. Enhancing the workflow with RAG prevents edits that worsen stronger requests, and it further amplifies improvements for weaker requests. Quality gains from RAG primarily come from the retriever agent selecting well-written examples for the LLM, as the quality of revisions declines only slightly when removing ratings from the explainer agent. Overall, using LLM revisions with RAG increases the predicted success rate for weaker requests by 14\% without degrading performance on stronger requests.
    
    The rest of this paper is organized into the following sections: \autoref{sec:lit-review} summarizes prior research and our contributions, \autoref{sec:text-data} describes text data on referral requests, \autoref{sec:measure-quality} measures the quality of a referral request by training models to predict its probability of receiving referrals, \autoref{sec:workflows} lays out workflows for improving referral requests with AI agents, \autoref{sec:workflow-results} reports on the effectiveness of LLM revisions, \autoref{sec:limitations} discusses key limitations, and \autoref{sec:future-experiments} concludes with a discussion on experiments that would address these limitations.
    
\section{Related Literature} \label{sec:lit-review}
    
    We highlight our contributions to three closest strands of literature: job referrals, making effective requests, and LLM performance on writing tasks. 
    
    \subsection{Job Referrals}
    
    Our study examines job referrals facilitated through anonymous online platforms, which differs from prior research focusing on offline interactions. For instance, \citet{beaman2012} and \citet{hensvik2016} conducted field experiments in developing countries to investigate how social networks influence job referrals. These studies benefit from controlled environments but are often limited to small sample sizes. Larger-scale studies, such as \citet{ilyes2023} and \citet{hensvik2016}, utilize administrative data but lack details on the content of referral interactions. Our study is unique in examining referral requests in a scalable online setting involving thousands of users with detailed text data.
    
    \subsection{Making Effective Requests} \label{subsec:effective-requests}
    
    While there is a well-developed literature on making effective requests in management and social psychology, our study is the first to specifically focus on requests for job referrals. Studies on online communities, such as the ``Random Acts of Pizza'' on Reddit \citep{althoff2014, majumdar2018}, suggest that attributes like gratitude, urgency, and reciprocity contribute to successful requests. Similarly, \citet{desai2015, yilmaz2022, herzenstein2011} emphasize the role of emotional appeal, narrative framing, and personal storytelling in driving successful requests in online contexts. These studies motivate the rich set of semantic attributes in our featurized model to predict successful referral requests, and they also motivate our use of transformer-based representations that are less reliant on pre-defined attributes.

    \subsection{Performance of LLMs on Writing Tasks}
    
    A rapidly growing literature examines how LLM agents perform on writing tasks, and our study contributes to understanding their persuasive ability for altruistic requests. LLMs have made it easier to tailor cover letters and resumes to job postings, which led to increased adoption by job applicants \citep{galdin2025cheap, cui2025signal}. However, evidence is nuanced on the quality of writing generated by LLMs. \citet{Noy2023} and \citet{deygers2025} find that using LLMs can improve performance on writing tasks, but this can depend on whether the task is within current capabilities of LLMs \citep{dellacqua2023}. There is growing evidence on the persuasive ability of LLMs to change human beliefs on topics like social policy, vaccinations, and conspiracy theories \citep{bai2025,havin2025aichange,Karinshak2023,Costello2024}. Our paper contributes to this literature by examining whether LLMs can write persuasive requests for an altruistic favor: providing employee referrals.

\section{Text Data on Job Referral Requests} \label{sec:text-data}

     \begin{figure*}[t]
      \centering \includegraphics[width=0.53\linewidth]{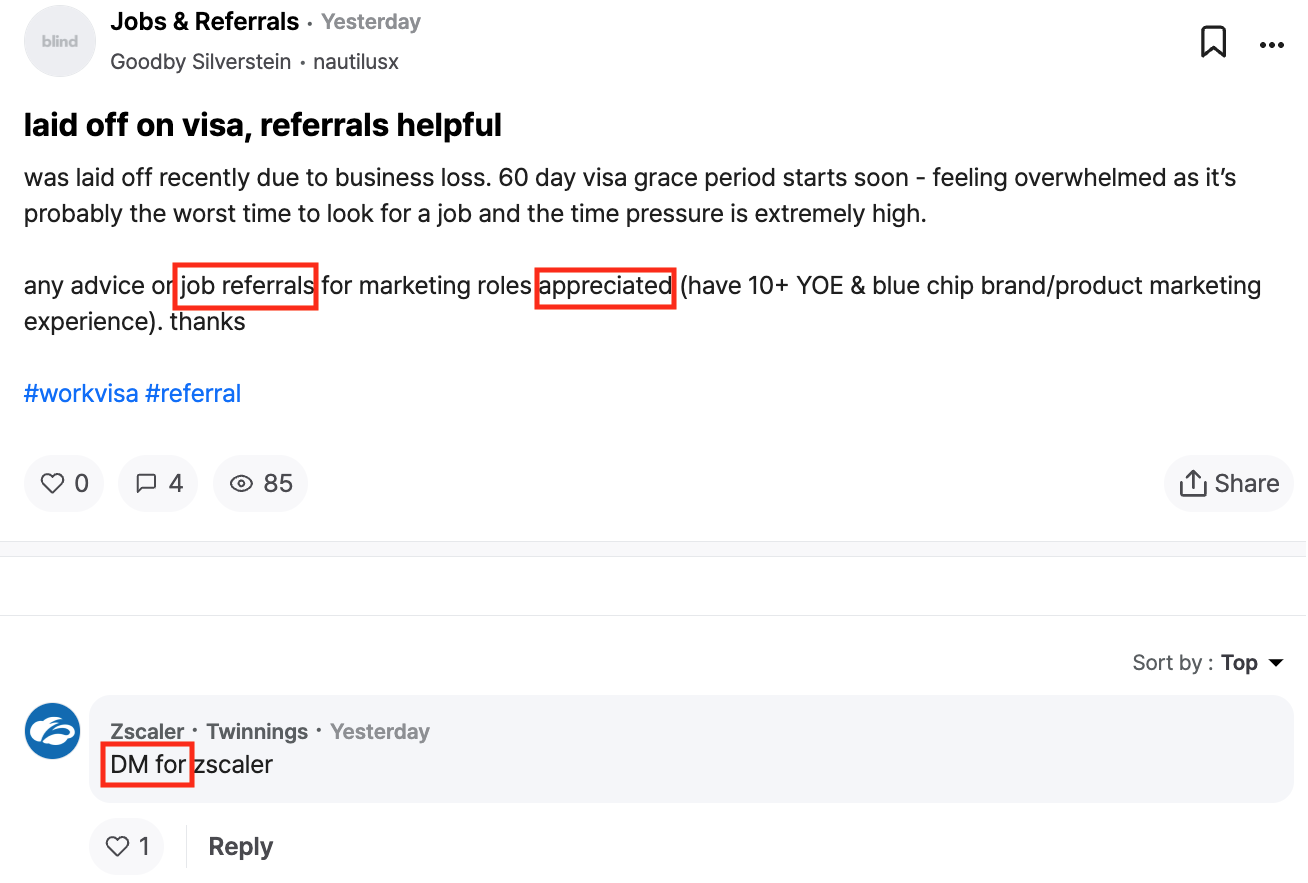}
      \caption{Example Referral Request on \texttt{teamblind.com}}
      \label{fig:teamblind}
    \end{figure*}
    
    \subsection{Background on the Blind Platform}
    
     Text data for our analysis come from Blind (\texttt{teamblind.com}), a professional online community for tech-sector employees. While this platform is not designed for any particular industry or job function, it primarily consists of employees in the technology sector. As detailed by \citet{chaudhary2023}, Blind is similar to Reddit in its anonymous discussion boards organized by topic channels, but a distinctive feature of Blind is that many users verify their accounts with corporate emails to access private channels for their companies. In our sample, 85\% of referral requests come from users with verified corporate credentials.
     
     We focus on the ``Jobs \& Referrals'' channel, where 80\% of posts are written by job seekers looking for employee referrals from other users. Both job seekers and employees benefit from these interactions. Job seekers improve their chances of securing interviews through referrals, while employees often receive cash rewards if a referred candidate is successfully hired. However, employees are selective about giving referrals because they face referral caps, reputation risks, and coordination costs in terms of time and effort. AI agents developed in this project create value to job seekers by helping them craft effective requests for referrals, and they also create value for employees by making it easier to identify promising candidates. 

    \subsection{Identifying Referral Requests and Offers}
    
     Text data from 17,542 posts and 27,150 comments have been collected from the ``Jobs \& Referrals'' channel between February 29th and November 17th, 2024. For each post, there is information on the date, title, content, username, and corporate affiliation, along with engagement metrics such as views, likes, and comments. For each comment, we collected data on text content, username, affiliation, and likes.
    
     \autoref{fig:teamblind} is a screenshot for an example referral request. We identified referral requests by looking for posts with generic terms and hashtags related to referrals (e.g. refer, referral, \#referral, \#needajob). Among posts containing these terms, we further narrowed them down to explicit requests for referrals, which contain phrases like "need a referral", "can anyone refer me", or "seeking a referral". This helps distinguish general discussions on referrals apart from direct requests for employee referrals. 
        
     In this channel, it is customary for users to offer referrals by asking the poster to contact them via direct messaging. We treat these comments as indicators of receiving job referrals, but a key limitation is that we cannot verify whether they lead to actual referrals because we do not observe activity outside the platform. 
    
     We identified referral offers by matching on phrases for direct messages and offers to help (e.g. ``DM me for Google'', ``happy to help''). Within this subset, we excluded comments that resemble requests more than offers (e.g. ``Can I also DM for Google?''). This step is necessary because posts with referral offers tend to attract comments by other job seekers who are also looking for referrals.
    
    \subsection{Concealing Credentials with Mask Tokens}
    
     Mask tokens were used to conceal certain credentials from all referral requests, such as profession, prior salary, seniority, years of experience, and current employer. This is to focus the model on \underline{how} the request is written rather than focusing on inherent advantages held by some job seekers. This also prevents AI agents from improving referral requests by exploiting attributes that are irrelevant to the job seeker. Tokenizers were updated so that the model can recognize these mask tokens (e.g. \texttt{[ROLE]}, \texttt{[LOCATION]}). 

    \begin{figure*}[t]
      \centering \includegraphics[width=0.5\linewidth]{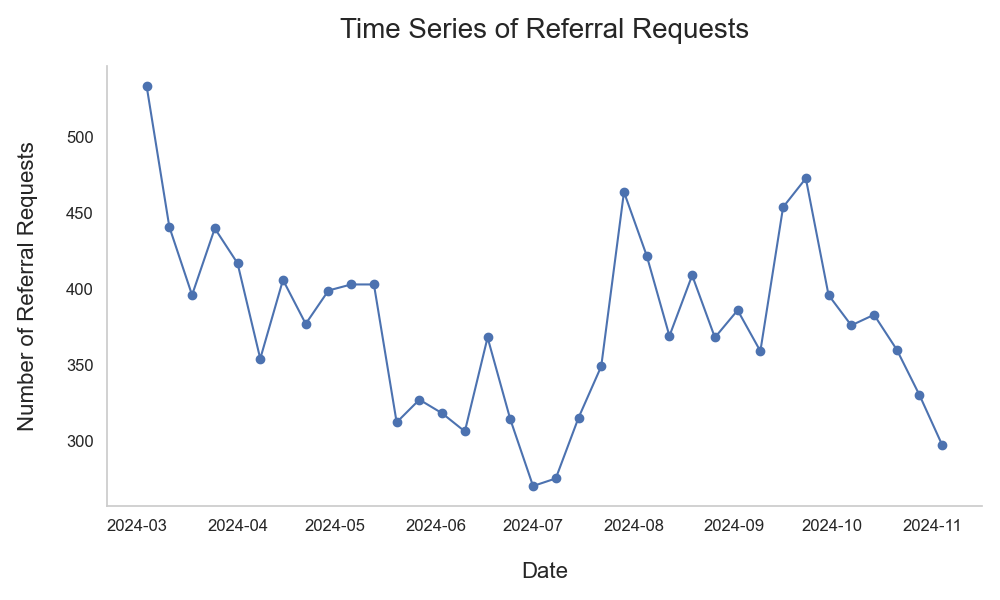}
    
      \includegraphics[width=0.32\linewidth]{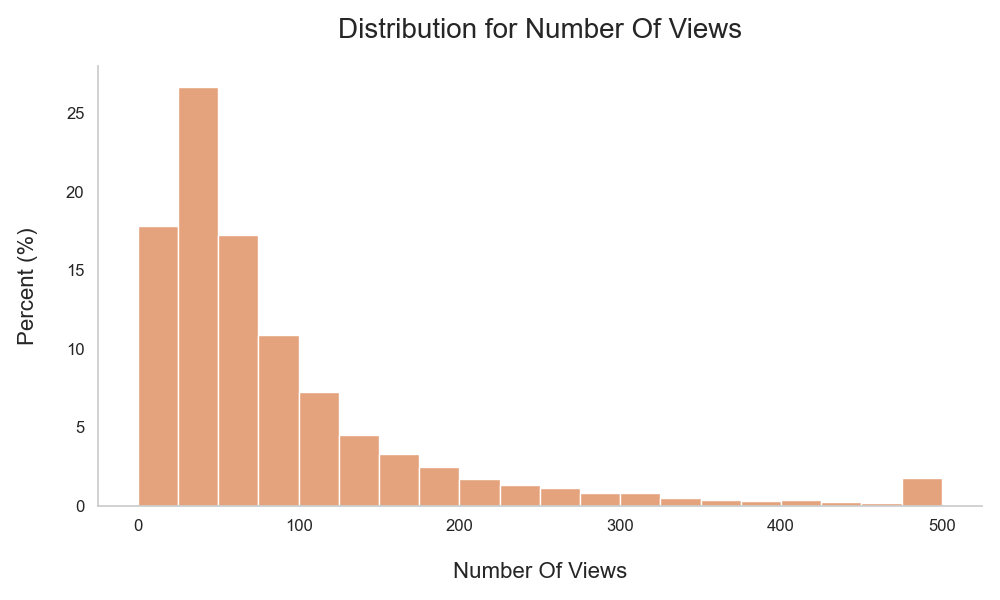} \hfill
      \includegraphics[width=0.32\linewidth]{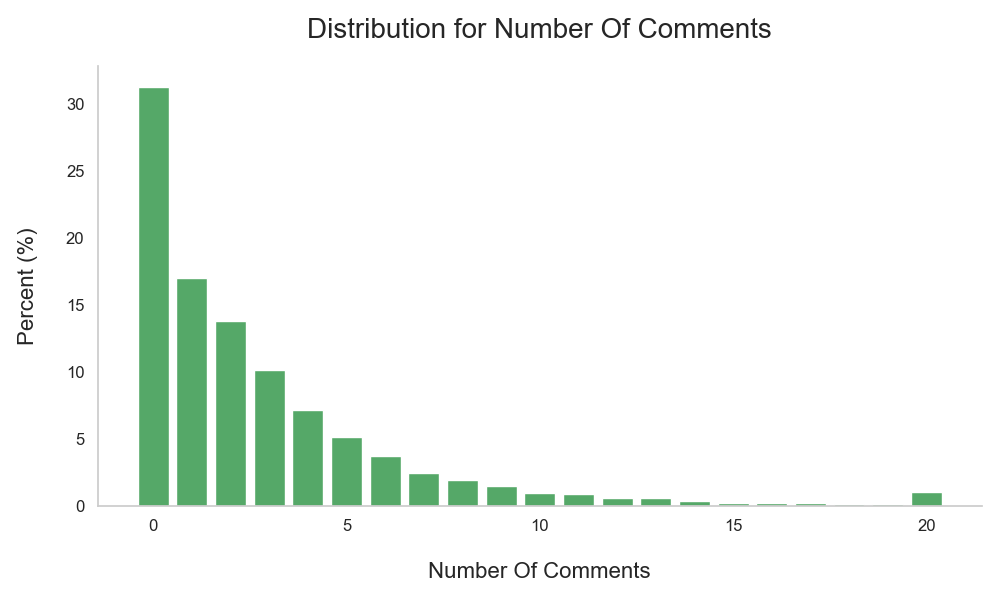} \hfill
      \includegraphics[width=0.32\linewidth]{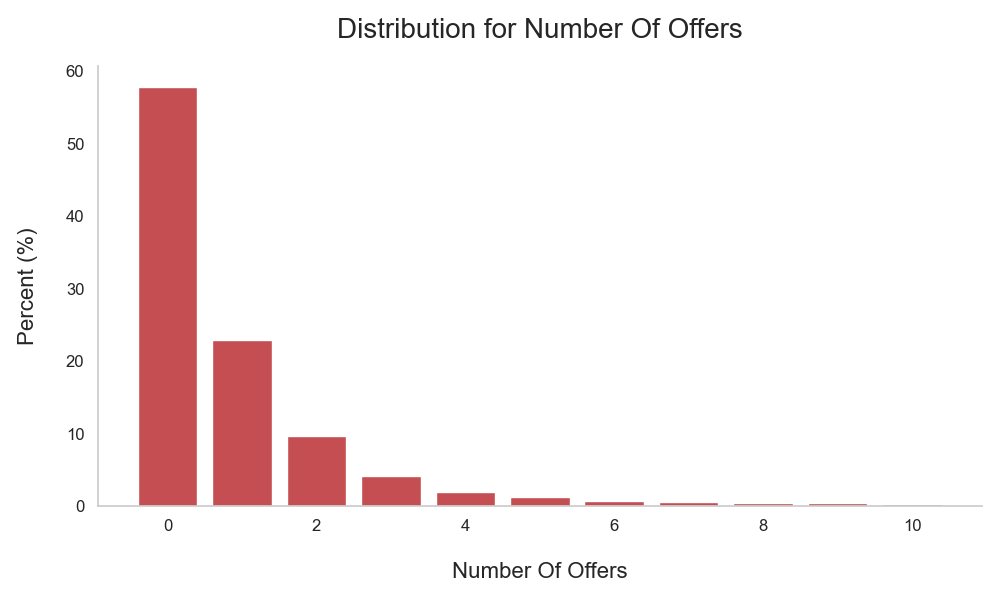}
      \caption{Descriptive Patterns of Referral Requests}
      \label{fig:descriptive}
    \end{figure*}

    \subsection{Descriptive Patterns}
    
     Empirical patterns suggest that activity on this platform reflects real job seeking behavior. During the 37-week sample period, 80\% of posts in this channel are referral requests, 85\% of which come from users with verified corporate credentials. The time series in \autoref{fig:descriptive} shows seasonal patterns in the volume of referral requests, which exhibit peaks and troughs that align with active recruitment cycles in the technology sector.
    
     Engagement metrics are highly skewed across referral requests. Histograms in \autoref{fig:descriptive} show the distribution of views, comments, and offers. Referral requests are mostly noticed by other users (views > 0), but many of them receive no engagement: 54\% of requests do not receive referral offers, and 31\% do not receive any comments at all. 

\section{Measuring the Quality of Requests} \label{sec:measure-quality}

     In the basic workflow for improving referral requests, an improver agent rewrites the referral request and an evaluator agent compares its quality before VS after LLM revisions. An important component of this workflow is defining a standard for the quality of referral requests. Following the approach in \citet{stiennon2020}, we measure the quality of a referral request by training a reward model to predict its probability of receiving referrals from other users. 

    \subsection{Prediction Task}
    
         We define a referral request to be "successful" if it receives any referral offers from other users. This is a classification task with text inputs and binary output labels. The input is a text string that contains the title and content of the request, along with mask tokens that obscure certain attributes (e.g. \texttt{[ROLE]}, \texttt{[LOCATION]}). The output is a binary label that indicates whether the request received at least one referral offer from other users. The objective is to train a model that predicts the probability of being successful given the textual content of the referral request. 

        \begin{table*}[h!]
            \centering
            \caption{Performance Metrics \& Bootstrapped Confidence Intervals}
            \label{tab:model_performance}
            \resizebox{\textwidth}{!}{%
            \begin{tabular}{@{}lccccc@{}}
            \toprule[2pt]
            \textbf{Model}          & \textbf{AUROC}        & \textbf{Accuracy}     & \textbf{Precision}    & \textbf{Recall}       & \textbf{F1-Score} \\
            \midrule
            Random Baseline         & 0.500                 & 0.499 (0.480-0.518)   & 0.473 (0.446-0.499)   & 0.459 (0.433-0.485)   & 0.466 (0.442-0.489) \\
            Featurized Model        & 0.576 (0.554–0.597)   & 0.562 (0.543–0.581)   & 0.545 (0.516–0.573)   & 0.472 (0.447–0.501)   & 0.506 (0.481–0.531) \\
            TF-IDF Model            & 0.647 (0.627–0.667)   & 0.602 (0.585–0.619)   & 0.586 (0.558–0.612)   & 0.556 (0.527–0.581)   & 0.570 (0.548–0.593) \\
            Sentence Transformer    & 0.681 (0.662–0.699)   & 0.630 (0.613–0.647)   & 0.613 (0.586–0.638)   & 0.604 (0.577–0.630)   & 0.608 (0.586–0.629) \\ 
            \bottomrule[2pt]
            \end{tabular}%
            }
        \end{table*}

        \begin{figure*}[htbp]
          \includegraphics[width=0.49\linewidth]{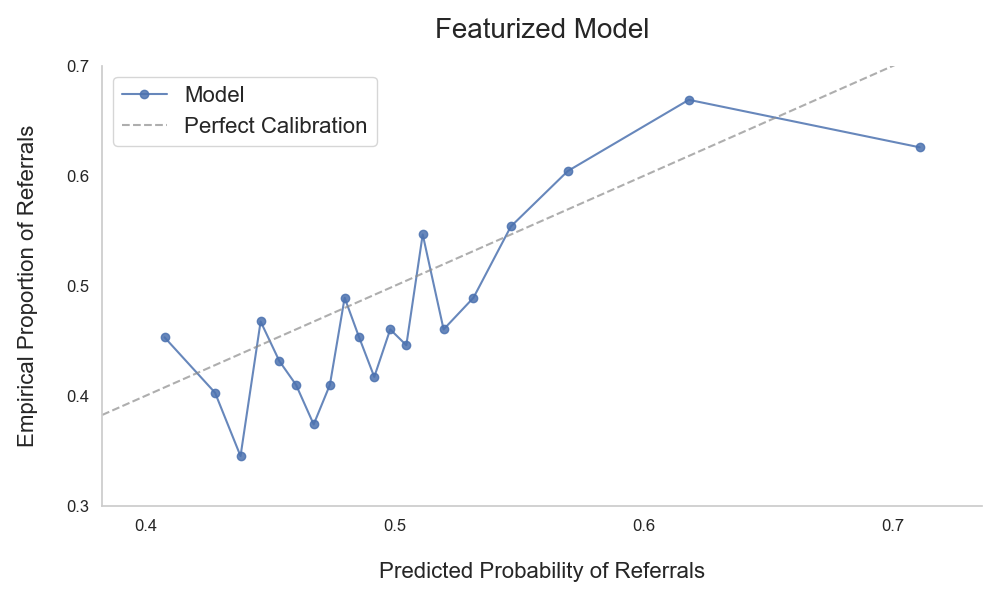} \hfill
          \includegraphics[width=0.49\linewidth]{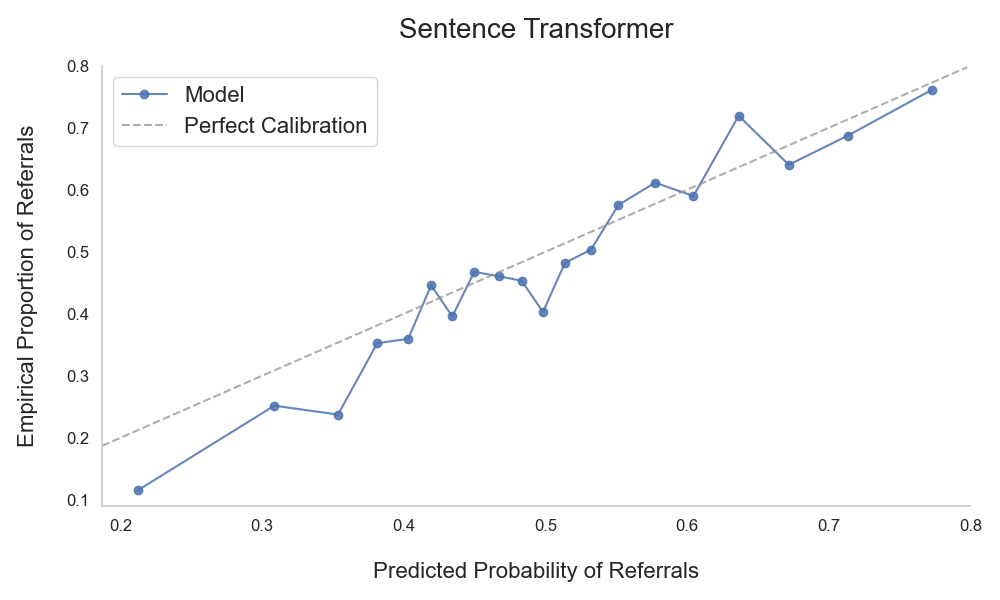}
          \caption{Predicted Probabilities VS Empirical Proportions Receiving Referrals}
          \label{fig:binscatter}
        \end{figure*}

    \subsection{Text Classification Models}
            
         We considered three models for encoding text from referral requests: 1) a TF-IDF model, 2) a sentence transformer, and 3) a featurized model with semantic attributes. We used a logistic regression model for the classification head, which includes a L1 penalty term selected on AUROC using grid search with cross validation on training data.
        
         The TF-IDF model (term-frequency inverse document frequency) encodes text by counting the frequency of unigrams and bigrams, and it scales by their overall frequency in training data. We removed stopwords and lemmatized tokens into their root forms. 
        
         The sentence transformer encodes the text string into a 768-dimensional vector. We used the \texttt{all-distilroberta-v1} model with 82M parameters, which was suitable for training and inference on our local machines. The model is pre-trained on contrastive learning with 1B sentence pairs, but we additionally fine-tuned parameters on the referral classification task using Low-Rank Adaptation (LoRA).\footnote{The best checkpoint was selected on AUROC.}
         
         The featurized model contains semantic attributes motivated from our domain knowledge of the platform  and prior literature on making effective requests. These features are binary indicators that equal one when a matching pattern has been found. Based on our browsing of the platform, \autoref{tab:teamblind_features} contains attributes that appear to be indicative of successful requests. Each feature in \autoref{tab:domain_features} is motivated from prior research, and we curated regular expressions that match descriptions in the second column. Additionally, we included characterizations of linguistic properties for each referal request: lexical diversity (type-token ratios), total word counts, readability scores, and spelling errors. 
    
    \subsection{Model Evaluation}
    
         To avoid inflating performance due to leakage, we split our data into a 80\% training set and 20\% testing set using a date threshold. All models were trained and fine-tuned only using referral requests prior to September 24th, and out-of-sample performance was evaluated on referral requests after September 25th. Since the platform evolves over time, this is an appropriate comparison for assessing the temporal validity of models trained on historical data. 
        
         We evaluate each model on five performance metrics: accuracy, precision, recall, F1 score, and area under the receiver operating characteristic curve (AUROC). These metrics provide a comprehensive view of classification performance, balancing both prediction correctness and the ability to identify true positives and negatives. We compare model performance against a random classifier benchmark, which is helpful for assessing whether predicting referrals is a learnable task. This benchmark randomly assigns a positive label with probability 0.462, the empirical rate of success during the training period.
    
    \subsection{Model Performance}
    
         \autoref{tab:model_performance} shows that predicting referrals is indeed a learnable task. This table reports performance metrics for each model and their bootstrapped confidence intervals in parentheses. All models outperform the random baseline in all classification metrics, and this level of performance is comparable to altruistic requests studied in \citet{althoff2014}.

         The best performing model in \autoref{tab:model_performance} is the sentence transformer, which achieves an AUROC of 0.681 and 63\% accuracy. This AUROC is 10 points above the featurized model and 3 points above the TF-IDF model, and the accuracy is 7 points above the featurized model and 3 points above the TF-IDF model. These improvements are somewhat expected, as sentence transformers can account for richer context between tokens that semantic features or simple token counts are unable to capture.
        
         The sentence transformer has the best alignment between predicted probabilities and empirical success rates. Panels in \autoref{fig:binscatter} are calibration plots that group referral requests into quantile bins by their predicted success rates, and each scatter point is the share of successful requests in that bin (y-axis) and their average predicted probabilities (x-axis). For both models, there is a positive relationship between predicted success rates and actual success rates. However, scatter points for the transformer are closer to the 45-degree line than the featurized model. This implies that whenever the transformer predicts a 37\% probability of receiving referrals, roughly 37\% of those requests actually turn out to receive referrals. These patterns suggest that the transformer is better at quantifying uncertainty in a way that closely aligns with empirical success rates. 

    \subsection{Distribution of Predicted Success Rates}

\begin{figure}[htbp]
          \centering \includegraphics[width=\linewidth]{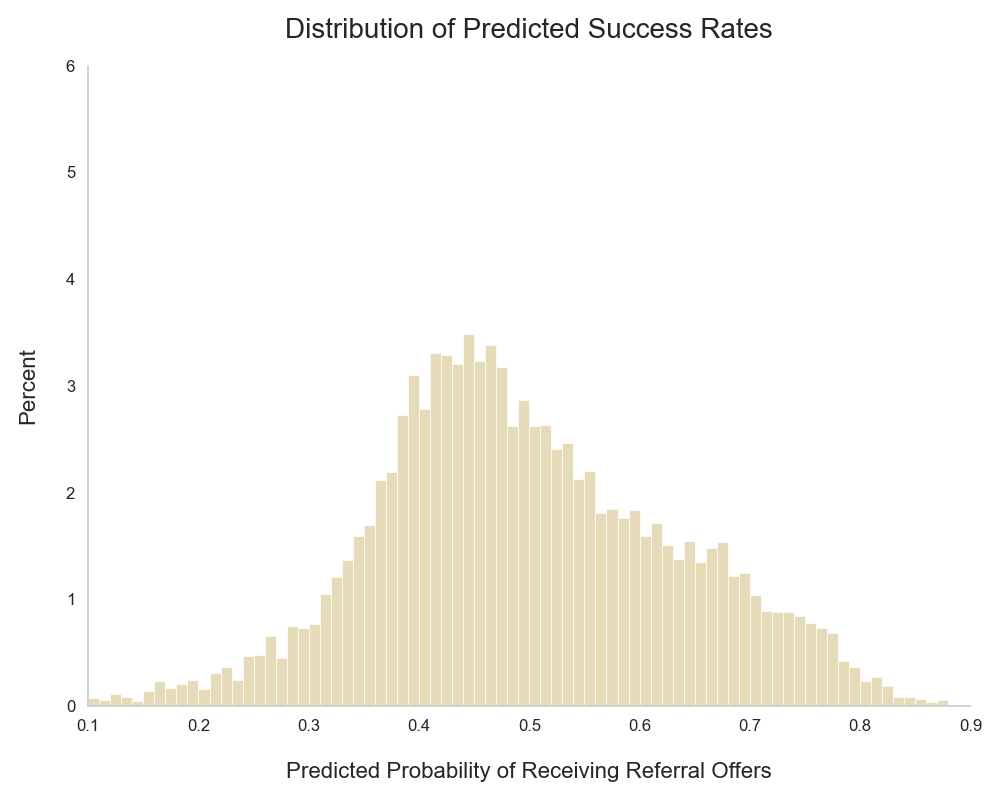} \hfill
          \caption{Distribution of Predicted Success Rates}
          \label{fig:prob_dist}
        \end{figure}

         The transformer predicts a wide range of success rates across referral requests. \autoref{fig:prob_dist} summarizes the distribution of predicted probabilities for receiving referrals, and \autoref{tab:prob_stats} reports its mean, standard deviation, and selected percentiles. Because mask tokens hide inherent advantages held by job seekers, predicted success rates mainly reflect the quality of writing in referral requests: some are well written and will likely receive referrals, while others have more room for improvement given their low success rates. 

        \begin{table}[htbp]
            \centering
            \caption{Distribution of Predicted Success Rates}
            \label{tab:prob_stats}
            \resizebox{\linewidth}{!}{%
            \begin{tabular}{lclclc}
            \toprule[2pt]
            Mean     & 0.497        & 1st pctile   & 0.182   & 75th pctile   & 0.590 \\
            Std Dev  & 0.136        & 5th pctile   & 0.287   & 90th pctile   & 0.688 \\
            Minimum  & 0.052        & 10th pctile   & 0.338   & 95th pctile   & 0.737 \\
            Maximum  & 0.881        & 25th pctile   & 0.404   & 99th pctile   & 0.801 \\
            Count    & N = 11,358   & 50th pctile   & 0.481   &    &  \\ 
            \bottomrule[2pt]
            \end{tabular}%
            }
        \end{table}
        
\section{Workflows for Improving Requests} \label{sec:workflows}

    \begin{figure*}[t]
      \centering \includegraphics[width=0.8\linewidth]{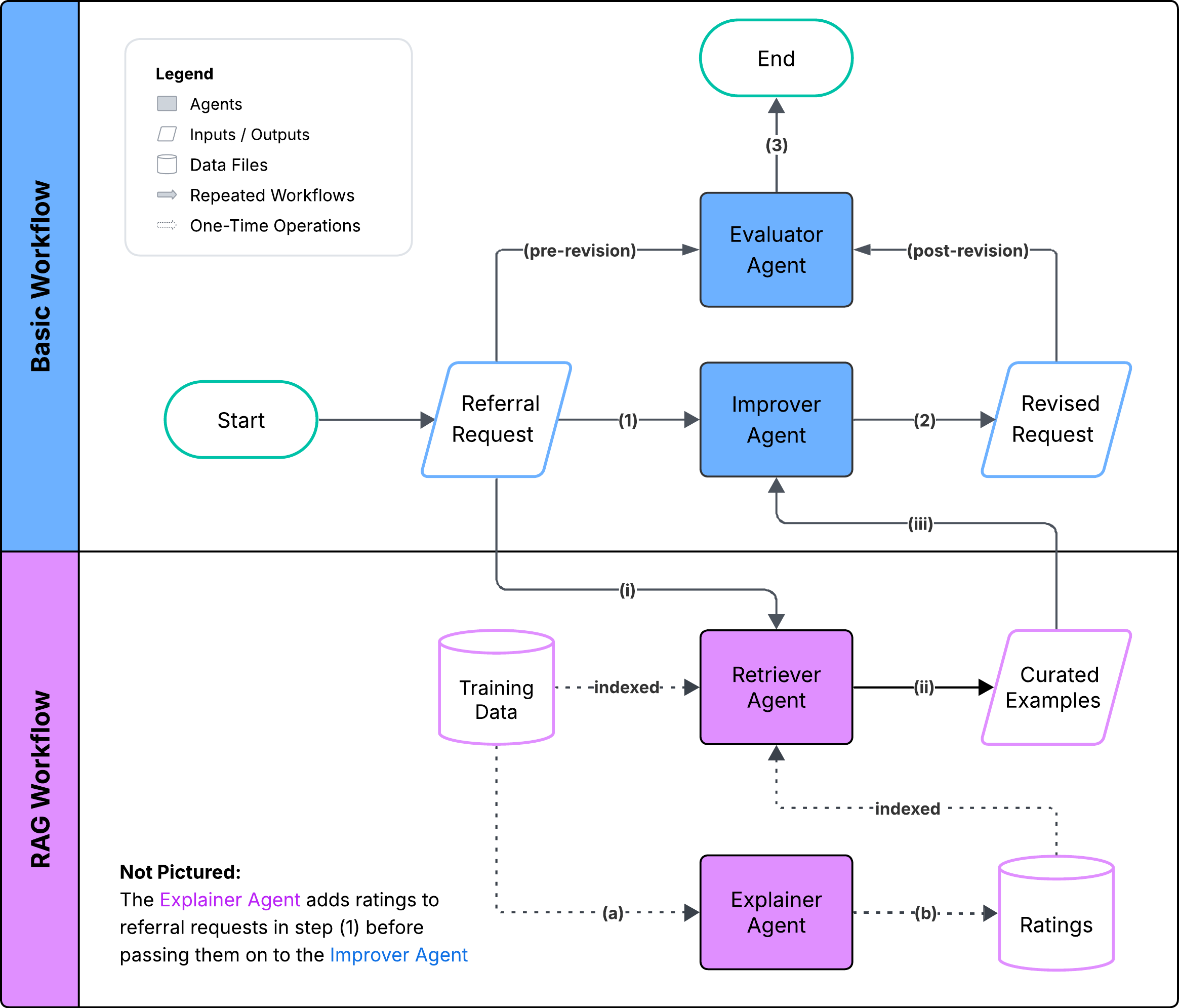} \hfill
      \caption{Workflows to Improve Requests for Job Referrals}
      \label{fig:diagram}
    \end{figure*}

    \subsection{Basic Workflow} \label{subsec:basic-workflow}

    The upper lane of \autoref{fig:diagram} shows the basic workflow to improve referral requests. The improver agent rewrites the referral request, and the evaluator agent compares the quality of the referral requests before VS after revisions suggested by the LLM.
     
    \subsubsection*{Improver Agent to Rewrite Referral Requests}

        The basic workflow starts in step (1), where the improver agent makes an API call to the \texttt{gpt-5-mini} model to rewrite the referral request. The LLM returns a new title and content for the revised request in step (2), which is given to the evaluator agent to be assessed for its quality. The system prompt provides general guidelines for rewriting the referral request to maximize its chances of receiving referrals, which includes well-written examples curated by the retriever agent. The user prompt is a structured JSON with metadata on the referral request, which includes the title, content, and quality ratings provided by the explainer agent (discussed in \autoref{subsec:rag-workflow}). For both the title and content, certain qualifications are intentionally concealed with mask tokens (e.g. \texttt{[ROLE]}, \texttt{[LOCATION]}) to prevent the LLM from exploiting attributes that are irrelevant to the job seeker. The prompt discourages the LLM from inventing false facts or injecting irrelevant attributes, but it is allowed to use mask tokens to include necessary qualifications (e.g. "I am a \texttt{[ROLE]} in \texttt{[LOCATION]}"). 

    \subsubsection*{Evaluator Agent to Measure Revision Quality}
    
         The evaluator agent measures the effectiveness of revisions suggested by the LLM by comparing predicted success rates for the original request and its revised version. Following the approach in \citet{stiennon2020}, the transformer model in \autoref{sec:measure-quality} is a reward function that serves as a proxy for the true outcome of interest: whether a request can successfully attract referral offers. Throughout this paper, the core assumption is that requests are more likely to receive referrals on the Blind platform if the model predicts higher success rates. The right panel of \autoref{fig:binscatter} supports this claim, as referral requests with higher predicted probabilities are indeed more likely to receive referrals. Although we believe this is a reasonable proxy for measuring the quality of referral requests, we acknowledge the limitation that model-based measures of quality may not necessarily translate to more referrals in the real world.

    \subsection{RAG Workflow} \label{subsec:rag-workflow}

     The lower lane of \autoref{fig:diagram} shows how we enhance this workflow with Retrieval-Augmented Generation (RAG). The retriever agent curates well-written examples for the improver agent, and the explainer agent provides ratings on the overall request, its title, and each individual sentence.

    \subsubsection*{Retriever Agent to Curate Examples}
    
         The retriever agent augments the system prompt with well-written examples that are contextually relevant to the user’s request. Simply picking examples with the highest predicted success rates is not ideal, as the LLM might learn from polished but irrelevant examples (e.g. user is a single male in his 20s, but retrieved examples mention children and a laid-off husband). The retrieval process ensures high-quality examples by selectively indexing successful requests and restricting retrieval to examples with higher predicted probabilities. The resulting search pool is large enough for the retriever to select examples that are contextually relevant to the user's request. 
    
         The retrieval process adds three steps prior to the basic workflow. In step (i), the referral request is sent to the retriever agent. In step (ii), the agent predicts its probability of success $p$ using the fine-tuned transformer, computes its potential upside $(p_{max}-p)$, and narrows down the pool to examples whose predicted success rates exceed $p$ by at least half of this gap. From this filtered pool, the retriever selects five examples whose transformer embeddings are closest to the user’s request (based on cosine similarity). In step (iii), the curated examples are passed to the improver agent and inserted into the system prompt, along with explicit instructions to not copy irrelevant details.

         Prior to this workflow, the retriever agent performs a one-time operation to index transformer embeddings for successful requests in the training data. First, it removes outliers from the indexed pool by trimming the top and bottom 3\% of requests based on their predicted success rates. Then, it uses a \texttt{FAISS} indexer to group the top 10\% of remaining requests into clusters for efficient retrieval.

    \subsubsection*{Explainer Agent for Editorial Guidance}

          The explainer agent provides ratings on the overall request, its title, and each individual sentence. Ratings can take on three values: ``strong'', ``weak'', or ``moderate''. The rating on the overall request guides the LLM on how heavily the request should be edited, and title/sentence ratings guide the LLM on specific segments it should focus on. The overall rating on the request is based on its predicted success rate, whereas title and sentence ratings additionally depend on attribution scores derived from Integrated Gradients \citep{sundararajan2017}.
          
          The attribution score for a sentence measures its contribution to the predicted probability of receiving referrals. Initially, integrated gradients are used to compute attribution scores for each token and embedding dimension. These values are summed across embedding dimensions for token-level attributions, which are summed again across tokens to obtain sentence-level attributions. These values are normalized by the total sum, so that each value represents the share of total attributions assigned to that sentence. 

          To improve the interpretability of attribution scores, each sentence is labeled with a ``strong'', ``weak'', or ``moderate'' rating. This is determined by how its attribution score stacks up against other sentences within the same request, as well as by the overall rating of the request. Since the title typically receives the largest attribution due to its key role in capturing attention from other users, we evaluate its attribution share relative to other requests of comparable length.
          
          The system prompt directs the LLM to use the overall rating to determine how intensive the edits should be: extensive edits for ``weak'' requests, minimal edits for ``strong'' requests, and balanced editing for ``moderate'' requests. Similarly, title and sentence ratings guide decisions on which sections to keep and which to rewrite. Strong sentences should be kept since they are effective at attracting referral offers, whereas weak sentences should be replaced with better alternatives. Moderate sentences may be kept or rewritten depending on the overall rating and how well they fit in with remaining edits.

          All examples chosen through RAG are augmented with ratings from the explainer agent. This is mainly a one-time procedure: in step (a), the explainer agent receives referral requests contained in the training data. In step (b), the explainer agent assigns ratings to each request, which are then indexed by the retriever and included with chosen examples.

        \begin{table*}[htbp]
            \caption{Predicted Probability of Receiving Referrals Before VS After LLM Revisions}
            \label{tab:main-results}
            \resizebox{\linewidth}{!}{%
            \begin{tabular}{cccc|ccc|ccc}
                \toprule[2pt] %
                
                \multirow{2}{*}{\textbf{Revision Type}} %
                & \multicolumn{3}{c}{\makecell[c]{\textbf{Lower Half}\\\textbf{($p<$ median)}}} %
                & \multicolumn{3}{c}{\makecell[c]{\textbf{Upper Half}\\\textbf{($p \geq$ median)}}} %
                & \multicolumn{3}{c}{\textbf{Overall}} \\ %
                
                \cmidrule(r){2-4} \cmidrule(r){5-7} \cmidrule(r){8-10} %
                
                 & \boldmath{{$p$}} & \boldmath{{$\Delta p$}} & \multicolumn{1}{c}{\textbf{improved}} %
                 & \boldmath{{$p$}} & \boldmath{{$\Delta p$}} & \multicolumn{1}{c}{\textbf{improved}} %
                 & \boldmath{{$p$}} & \boldmath{{$\Delta p$}} & \multicolumn{1}{c}{\textbf{improved}} \\ %
                 
                \hhline{|----------|} %
                 
                Original Request %
                & \makecell[c]{0.392\\(0.002)} & & %
                & \makecell[c]{0.607\\(0.002)} & & %
                & \makecell[c]{0.499\\(0.003)} & & \\ %
                
                Basic Workflow %
                & \makecell[c]{0.426\\(0.003)} & \makecell[c]{0.034\\(0.003)} & \makecell[c]{0.637\\(0.013)} %
                & \makecell[c]{0.572\\(0.003)} & \makecell[c]{-0.034\\(0.003)} & \makecell[c]{0.363\\(0.013)} %
                & \makecell[c]{0.499\\(0.003)} & \makecell[c]{-0.000\\(0.002)} & \makecell[c]{0.500\\(0.009)} \\ %
                
                RAG Workflow %
                & \makecell[c]{0.447\\(0.003)} & \makecell[c]{0.055\\(0.003)} & \makecell[c]{0.706\\(0.012)} %
                & \makecell[c]{0.608\\(0.003)} & \makecell[c]{0.002\\(0.003)} & \makecell[c]{0.509\\(0.013)} %
                & \makecell[c]{0.527\\(0.003)} & \makecell[c]{0.028\\(0.002)} & \makecell[c]{0.607\\(0.009)} \\ %
                
                Exclude Ratings %
                & \makecell[c]{0.446\\(0.003)} & \makecell[c]{0.054\\(0.003)} & \makecell[c]{0.700\\(0.012)} %
                & \makecell[c]{0.604\\(0.003)} & \makecell[c]{-0.002\\(0.003)} & \makecell[c]{0.497\\(0.013)} %
                & \makecell[c]{0.525\\(0.003)} & \makecell[c]{0.026\\(0.002)} & \makecell[c]{0.599\\(0.009)} \\ %
        
            \bottomrule[2pt]
            \end{tabular}%
            }
            \, \\ \, \\
            \footnotesize Standard errors are in parentheses. $p$ is the predicted success rate, $\Delta p$ is its change after LLM revisions, and ``improved'' is the share of requests with $\Delta p > 0$. Referral requests are grouped into upper and lower halves based on their predicted success rates prior to LLM revisions. The last row (``Exclude Ratings'') denotes the RAG workflow without ratings from the explainer agent.
        \end{table*}

        \begin{figure*}[t]
          {
          \centering 
          \includegraphics[width=0.49\linewidth]{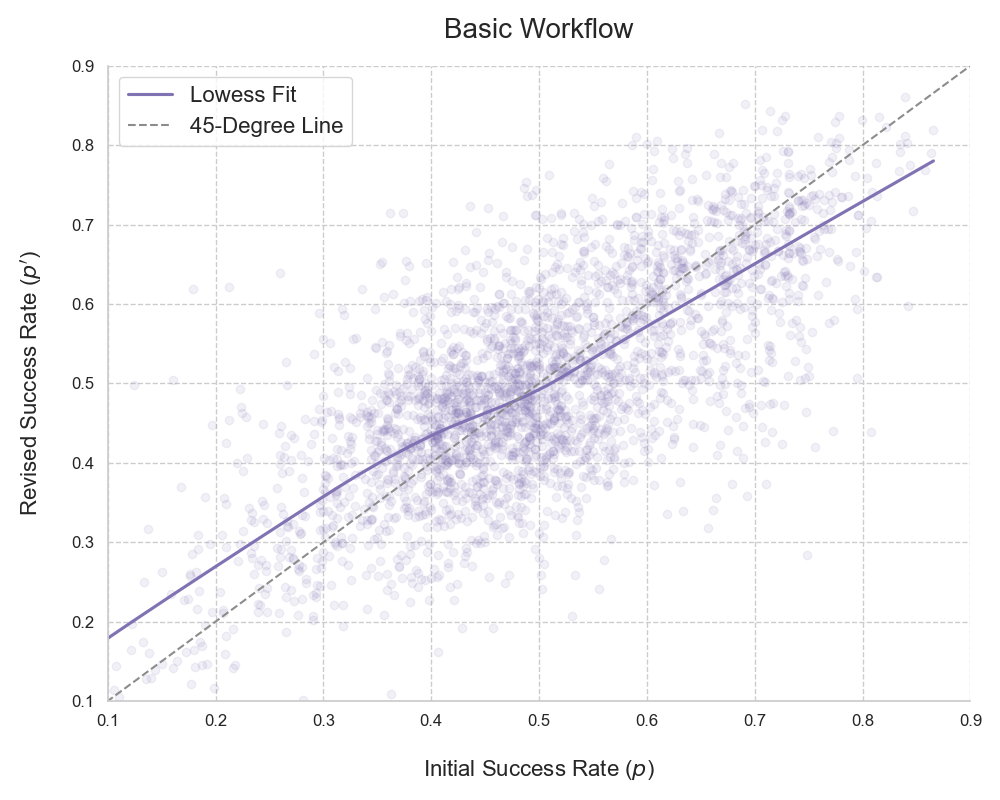} \hfill
          \includegraphics[width=0.49\linewidth]{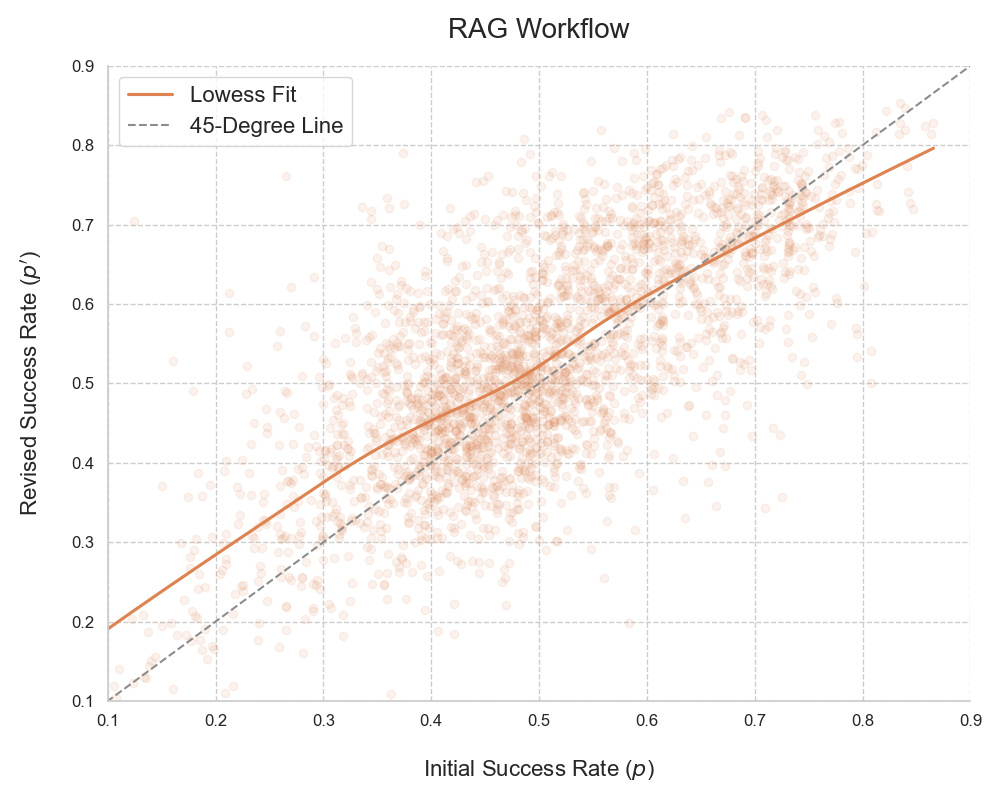} 
          \caption{Predicted Probability of Receiving Referrals Before VS After LLM Revisions}
          \label{fig:workflow-calib}
          }
          \, \\
          \footnotesize Each scatter point is a referral request, and axes indicate predicted probabilities of receiving referrals before VS after LLM revisions. Scatter points are above the 45-degree line if predicted success rates improve after LLM revisions, and below the 45-degree line if they worsen after LLM revisions. The solid line represents the nonparametric Lowess curve for the smoothed average of post-revision success rates for a local bandwidth of initial success rates.
        \end{figure*}

\section{Effectiveness of LLM Revisions} 
\label{sec:workflow-results}

    Revisions proposed by the LLM affect the quality of referral requests in an asymmetric manner, which varies according to the initial probability of receiving referrals. The basic workflow increases predicted success rates for weaker requests but decreases them for stronger ones. Using RAG to enhance the workflow avoids changes that degrade already strong requests while amplifying improvements for weaker requests. Most of the quality improvement from RAG come from the retriever agent choosing well-written examples for the LLM; removing ratings from the explainer agent only slightly reduces the quality of the revisions. Overall, using LLM revisions with RAG increases the predicted success rate for weaker requests by 14\% without degrading performance on stronger requests.

    \subsection{Asymmetric Impact of the Basic Workflow}

        The basic workflow increases predicted success rates for weaker requests but decreases them for stronger requests. This can be seen in the left panel of \autoref{fig:workflow-calib} and the first two rows of \autoref{tab:main-results}. Scatter points in \autoref{fig:workflow-calib} compare predicted success rates before VS after LLM revisions, and the solid line is a Lowess curve representing the smoothed average of post-revision success rates along initial probabilities of receiving referrals. \autoref{tab:main-results} summarizes the effectiveness of LLM revisions for stronger and weaker requests, defined by whether their initial success rates $p$ are above or below the median. 
        
        For weaker requests, average success rates are \underline{higher} by $\Delta p$ = 3.4 percentage points in the basic workflow, which is 8.6\% above the average for original requests ($p$ = 0.392). For stronger requests, average success rates are \underline{lower} by $\Delta p$ = -3.4 percentage points, which is -5.7\% below the average for original requests (avg $p$ = 0.607). \autoref{fig:workflow-calib} shows similar patterns, where the Lowess curve is above the 45-degree line for lower values of initial probability $p$ and below the 45-degree line for higher values of initial $p$. 64\% of requests in the lower half have $\Delta p >0$ (improved success rates), while the corresponding share is only 36\% for stronger requests in the upper half of initial $p$.

        This asymmetric effect is analogous to what \citet{Brynjolfsson2025} find for customer support agents: AI assistance enhances output quality for less experienced workers and slightly diminishes quality for more experienced workers. Although reported magnitudes for quality degradation are larger in the above results, they become more comparable once the workflow is enhanced with RAG, which is better at mitigating edits that worsen the quality of referral requests. 
        
    \subsection{Improvements with the RAG Workflow}
    
        Enhancing the workflow with RAG prevents edits that worsen stronger requests, and it further amplifies improvements for weaker requests. This can be seen in the right panel of \autoref{fig:workflow-calib} and the third row of \autoref{tab:main-results}. For weaker requests, RAG increases $\Delta p$ from +3.4 percentage points to +5.5 percentage points, corresponding to an improvement that is 14\% higher than the average probability observed for the original requests ($p$ = 0.392 $\rightarrow$ 0.447). RAG reduces $\Delta p$ from -3.4 percentage points to +0.2 percentage points, which is no longer statistically distinguishable from the original request. The right panel of \autoref{fig:workflow-calib} shows similar patterns, as the Lowess curve is further above the 45-degree line for lower values of initial $p$ while being closer to this line for higher values of initial $p$.
        
\begin{figure}[t]
      \centering
      \includegraphics[width=\linewidth]{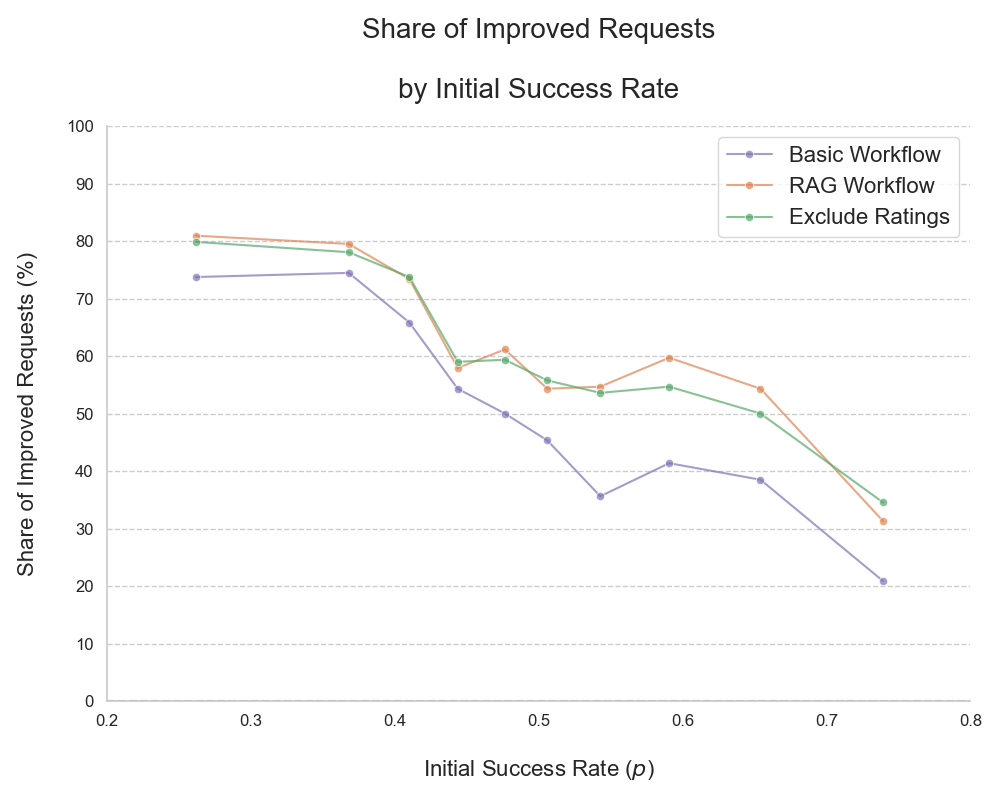}
      \caption{Share of Improved Requests by Initial Quality}
      \label{fig:share-improved}
    \end{figure}

        \autoref{fig:share-improved} shows that the RAG workflow mainly increases the share of improved outcomes for stronger requests. This figure plots the share of requests with $\Delta p > 0$ by deciles of initial success rate $p$. The gap between purple and orange scatter points shows the difference in the proportion of improved requests, which grows substantially for larger initial $p$. In the upper half, the share of improved requests increases by 15 percentage points when the basic workflow is augmented with RAG (from 36\% to 51\%), while the corresponding gain is only 7 percentage points in the lower half of requests (from 64\% to 71\%).

    \subsection{Factors Contributing to Gains from RAG}
    
        Examples retrieved by RAG support the LLM in two ways. First, the LLM can draw on well-written examples to improve weaker requests. Second, it can recognize stronger requests and intervene less, avoiding revisions that would lower their predicted success rates. One possibility is that the LLM primarily mimics high-quality examples: it can lift weaker requests toward the level of good examples but cannot reliably generate content that pushes beyond the existing upper bound on predicted success rates. This would explain why RAG helps prevent the LLM from worsening strong requests but does not meaningfully improve them.

        Quality gains achieved through RAG primarily come from examples curated by the retriever agent rather than from ratings provided by the explainer agent. The last two rows of \autoref{tab:main-results} only show small declines in the quality of revisions after removing ratings from the explainer agent. These differences are not statistically significant. The share of improved requests in \autoref{fig:share-improved} is also similar for the RAG workflow with and without ratings from the explainer agent (green VS orange scatter points). This minor decrease in quality suggests that the LLM learns more from well-written examples than from prescriptive guidance on specific segments of the referral request.

\section{Limitations} \label{sec:limitations}

    Due to important limitations, these workflows should mostly be viewed as proof-of-concept prototypes. Future work would need to address two major issues: limited external validity and reliance on model-based measures of success.

    First, the external validity of this analysis is limited. This paper focuses on the Blind platform, which evolves over time and is heavily concentrated in the tech sector. As a result, it is unclear whether these findings generalize to other platforms, labor markets, or time periods. Because we do not conduct human evaluations, we also cannot be certain that revised requests are in fact more appealing to human readers than their original counterparts. Moreover, it is unknown how these results would scale if many users adopted the tool, which could shift platform norms in ways that make suggested revisions less effective.

    Second, improvements in model-predicted success rates do not guarantee better outcomes in the real world. LLM revisions increase predicted success according to the reward model, which is informative but not perfect at predicting referral offers. Furthermore, we cannot observe whether commenters who offer referrals actually follow through with job seekers outside the platform, so realized referrals remain unmeasured. Field experiments that track real outcomes for job seekers were beyond the scope of this project, given operational and resource constraints they would impose on the platform. As a result, model-defined gains may not fully align with real-world improvements, and RAG may amplify idiosyncrasies of the reward model by preferentially retrieving examples favored by that model. These are inherent limitations of model-based proxy measures, although we mitigate the most severe forms of Goodhart’s law by not directly optimizing on the reward model.

\section{Conclusion} \label{sec:future-experiments}

    This paper develops AI agents that help job seekers write more effective requests for job referrals. Our approach has three components: (1) training a reward model to predict the probability of receiving referrals, (2) building a RAG system to guide the LLM with well-written examples, and (3) evaluating the impact of LLM revisions on the quality of referral requests. The basic workflow increases predicted success rates for weaker requests but tends to reduce them for stronger ones. Augmenting this workflow with RAG avoids edits that worsen stronger requests while further amplifying gains for weaker requests.

    Although increases in model-predicted success rates do not guarantee more referrals in the real world, they provide low-cost signals for prioritizing promising features before running higher-stakes experiments on real users. Online A/B tests can measure whether improvements in predicted success rates translate into more referrals for job seekers. Researchers can follow up with a subsample of job seekers to assess how often referral offers in the comments lead to actual referrals. Researchers can also track how often LLM revisions are adopted by job seekers, which validates their usefulness in the real world. Even if some job seekers reject suggested revisions, their causal impact can be estimated with experimental designs that account for imperfect compliance. Overall, model-based quality measures can help identify promising features before experimental validation in the real-world.

\bibliography{custom}

\appendix
\renewcommand{\thefigure}{A\arabic{figure}}
\renewcommand{\thetable}{A\arabic{table}}
\setcounter{figure}{0}
\setcounter{table}{0}

\clearpage

\vfill

\begin{table*}[h]
    \textbf{\raggedright \Large Appendix Tables \\ \, \\ \, \\ \, \\ \,}
    \centering
    \caption{Semantic Attributes Motivated from Platform-Specific Domain Knowledge \\ \,}
    \label{tab:teamblind_features}
    \resizebox{\textwidth}{!}{%
    \begin{tabular}{@{}p{3cm}p{7cm}p{6cm}@{}}
    \toprule[2pt]
    \textbf{Attribute} & \textbf{Description} & \textbf{Example Snippet} \\ 
    \midrule
    Mentions Layoff & Mentions layoffs using regex detection & \textit{"I was laid off last week."} \\
    Mentions PIP & Mentions performance improvement plan & \textit{"I might get PIP'd soon."} \\
    Mentions Company & Detects presence of company names. & \textit{"Discussing [FIRM\_NAME]."} \\
    Mentions Job Title & Regex detection of specific job titles. & \textit{"Seeking software engineering roles."} \\
    Year of Experience & Mentions years of experience using regex. & \textit{"10+ years of experience in tech."} \\
    Mentions Salary & Mentions salary or total compensation details. & \textit{"My TC is \$ 150k."} \\
    Reason for Search & Reasons like better balance or transition. & \textit{"Looking for better work-life balance."} \\
    Past Experience & Describes past roles or accomplishments. & \textit{"Worked as a consultant at X."} \\
    Mentions Skills & Detects mentions of specific skills. & \textit{"Skilled in Python and SQL."} \\
    \bottomrule[2pt]
    \end{tabular}%
    }
\end{table*}

\vfill

\clearpage

\begin{table*}
    \centering
    \caption{Semantic Attributes Motivated from Prior Literature}
    \label{tab:domain_features}
    \resizebox{\textwidth}{!}{%
    \begin{tabular}{@{}p{3cm}p{5cm}p{4cm}p{4cm}@{}}
    \toprule[2pt]
    \textbf{Attribute} & \textbf{Description}& \textbf{Example Snippet} & \textbf{Reference and Context} \\ 
    \midrule
    Urgency & Detects urgency in the text & \textit{"Need this job ASAP!"} & \citep{althoff2014}: Highlighted urgency as a motivator for action in altruistic requests. \\ 
    Gratitude & Detects expressions of gratitude. & \textit{"Thank you for the advice."} & \citep{althoff2014}: Found gratitude improves the likelihood of receiving help in altruistic contexts. \\ 
    Politeness & Detects politeness markers. & \textit{"Could you please help?"} & \citep{danescu2013}: Politeness fosters positive interpersonal impressions in communication. \\ 
    Familiarity & Informal or familiar greetings. & \textit{"Hey everyone, need help!"} & \citep{saunderson2021}: Familiar tone increases persuasiveness in human-robot interaction. \\ 
    Being Desperate & Identifies desperation in the text. & \textit{"I'm struggling, please help!"} & \citep{majumdar2018}: Emotional appeals rooted in desperation evoke sympathy. \\ 
    Inclusive/Exclusive & Tracks pronoun usage (we, they) & \textit{"We need to work together."} & \citep{saunderson2021}: Pronouns reflect social inclusion or individuality, influencing tone. \\ 
    Being Content & Detects positive sentiments about job satisfaction. & \textit{"Love my current job!"} & \citep{herzenstein2011}: Positive sentiment fosters trust and credibility in social interactions. \\ 
    Readiness & Mentions interview preparation. & \textit{"Practicing mock interviews."} & \citep{althoff2014}: Demonstrating preparation builds confidence in the requestor’s capability. \\ 
    Evidentiality & Claims expertise or experience. & \textit{"Have strong experience in data science."} & \citep{danescu2013}: Citing evidence reinforces credibility and persuasiveness. \\ 
    Reciprocity & Detects offers to help in return. & \textit{"Happy to assist in return."} & \citep{althoff2014}: Reciprocity strengthens social bonds and mutual obligations. \\ 
    High Status & Mentioning high-status phrases. & \textit{"Led teams for 10 years."} & \citep{herzenstein2011}: High-status indicators project authority and reliability. \\ 
    Gain/Loss Framing & Gain or loss framing in text. & \textit{"This job would change my life."} & \citep{majumdar2018}: Framing appeals highlight stakes to create urgency or optimism. \\ 
    \bottomrule[2pt]
    \end{tabular}%
    }
\end{table*}

\clearpage

\end{document}